\pdfoutput=1

\documentclass[11pt]{article}

\usepackage[]{acl}
\usepackage{booktabs}
\usepackage{graphicx}
\usepackage{multirow}
\usepackage{amsmath}
\usepackage{CJKutf8}
\usepackage[utf8]{inputenc}
\usepackage{booktabs}
\usepackage{multirow}
\usepackage{makecell}
\usepackage{amssymb}
\usepackage{pifont}
\usepackage{arydshln}
\usepackage{natbib}
\usepackage{color}
\usepackage{wasysym}
\usepackage{subcaption}
\usepackage{appendix}
\usepackage{float}
\usepackage{caption}
\usepackage{cuted}

\usepackage{times}
\usepackage{latexsym}

\usepackage[T1]{fontenc}

\usepackage[utf8]{inputenc}

\usepackage{microtype}

\title{Word Matters: What Influences Domain Adaptation in Summarization?}

\author{Yinghao Li$^{1}$\thanks{~~These authors are both First Author.}~~~Siyu Miao$^{1*}$~~~Heyan Huang$^{12}$\thanks{~~Corresponding author}~~~Yang Gao$^{12\dagger}$ \\
$^{1}$School of Computer Science and Technology, Beijing Institute of Technology, Beijing, China \\
$^{2}$Beijing Institute of Technology Southeast Academy of Information Technology, Putian, China \\
\texttt{\{yhli,symiao,hhy63,gyang\}@bit.edu.cn}
}

\begin{document}
\maketitle

\begin{abstract}

Domain adaptation aims to enable Large Language Models (LLMs) to generalize domain datasets unseen effectively during the training phase. However, factors such as the size of the model parameters and the scale of training data are general influencers and do not reflect the nuances of domain adaptation performance. 
This paper investigates the fine-grained factors affecting domain adaptation performance, analyzing the specific impact of `words' in training data on summarization tasks.
We propose quantifying dataset learning difficulty as the learning difficulty of generative summarization, which is determined by two indicators: word-based compression rate and abstraction level.
Our experiments conclude that, when considering dataset learning difficulty, the cross-domain overlap and the performance gain in summarization tasks exhibit an approximate linear relationship, which is not directly related to the number of words.
Based on this finding, predicting a model's performance on unknown domain datasets is possible without undergoing training.
Source code and scripts are available at \url{https://github.com/li-aolong/Word-Matters}.

\end{abstract}

\section{Introduction}

    With the continuous development of Large Language Models (LLMs), remarkable capabilities have been demonstrated in knowledge comprehension~\citep{thirunavukarasu2023trialling, sun2023headtotail}, logical reasoning~\citep{hao2023reasoning,miao2023selfcheck}, problem-solving~\citep{chan2023chateval,talebirad2023multi}, and other aspects~\citep{zhao2023survey,wen2023empowering}. 
    
    As a result, LLMs have been widely applied to various summarization tasks on different domains to improve productivity like law~\citep{shukla2022legal}, medicine~\citep{vanveen2023clinical}, finance~\citep{li2023large} and so on, including both natural and social science study~\citep{glickman2024ai,xu2024ai}.
    However, when LLMs are applied to specific domains, it often necessitates the selection of corresponding domain-specific knowledge bases for training~\citep{zhang2023fineval,cui2023chatlaw,thirunavukarasu2023large,yu-etal-2021-adaptsum}. This results in a limitation where a model trained in one domain struggles to be effectively applied in others~\citep{dada2023impact}, leading to a waste of resources. 
    
    This constraint arises from the disparity in the distribution between the training data and the target domain data~\citep{zhang2023nico++}. In light of this limitation, effective methods must be taken to fix the gap and then enhance the model's adaptability and efficiency in summarization tasks. Domain adaptation aims to train a model from multiple source domains, enabling it to generalize well to unseen domains~\citep{li2018deep, dou2019domain}. Consequently, enhancing domain adaptation performance is a key objective for large-scale models in improving downstream tasks~\citep{zhou2022domain}. It is worthwhile to explore which factors can affect the domain adaptation performance~\citep{Wang2021Generalizing}. 

    \citet{schaeffer2023emergent} proposes that metrics based on nonlinear or non-contiguous tokens are crucial to a model demonstrating emergent abilities and that ROUGE-L-Sum shows sharper variations. This has inspired us to consider the performance changes of models in domain adaptation from the perspective of more granular units. Tokens typically do not possess complete semantics, whereas words are the basic language units with specific meanings or functions. Therefore, we consider exploring the impact on model performance in domain adaptation from the perspective of words.

    Summarization tasks involve generating concise texts that encapsulate the main components of longer documents, considering factors such as coherence, information diversity, and coverage scope~\citep{ALOMARI2022101276}. This differs from other downstream tasks like machine translation~\citep{klimova2023neural} and classification~\citep{bird2023chatbot}. \citet{fatima2022Novel} note that reducing summary extractors' size or compression ratio can lead to losing vital content, features, concepts, and other significant information. Therefore, we explore how the degree of information extraction between input documents and target summaries impacts domain adaptation performance in summarization tasks.

This paper investigates how words impact the domain adaptation of summary tasks. 
We first introduce two indicators, compression rate, and abstraction level, to quantify the learning difficulty of datasets, thereby more accurately reflecting the performance gain of models. 
Then, we identify two key aspects affecting domain adaptation: cross-domain overlap and word count, hypothesizing a linear relationship between them and model performance.
Experiments are conducted with models of various sizes on summarization datasets from four domains. The results indicate that the cross-domain overlap exhibits an approximately linear relationship with performance gain when considering dataset learning difficulty. 
In contrast, word count shows no significant correlation. Based on this linear relationship, it is possible to predict model performance without undergoing training by using the cross-domain overlap calculated from the dataset. Our contributions can be summarized as follows:
    \begin{itemize}
    \item We propose two factors affecting the domain adaptation of summarization tasks: (1) \textbf{Learning difficulty coefficient} of the dataset more accurately reflects the performance gain; (2) \textbf{Cross-domain overlap} directly represents the closeness between the source and target domains.
    \item Our experiments show that cross-domain overlap has an approximately linear relationship with performance gains based on the learning difficulty coefficient, revealing the connection between datasets and domain adaptation from the perspective of words.
    \item We demonstrate that without undergoing training, it is possible to predict a model's performance on unknown domain datasets solely based on the learning difficulty coefficient and cross-domain overlap. This provides a resource-efficient and rapid validation method for models regarding domain adaptation.
    \end{itemize}

\section{Related Work}
Domain Adaptation (DA) has emerged as a crucial methodology for enhancing model performance across varying domains~\citep{farahani2020brief}. 
DA aims to enhance the performance of LLMs in a target domain, where annotated data may be scarce or absent, by leveraging knowledge from a related domain with a sufficient amount of labeled data~\citep{farahani2020brief}. 
Many methods have been developed to tackle the out-of-domain adaptation issue~\citep{zhou2021domain,fan2021adversarially,cha2021swad,wang2022generalizing,ling2023domain}. There are three pivotal strategies related to our work: (1) Continual pre-training, (2) Alignment of distributions, and (3) Adaptation tuning.  

\paragraph{Continual Pre-training}
Continual pre-training uses similar training objectives as continual self-supervised training to update pre-trained models with new data instead of retraining from scratch~\citep{gupta2023continual}. Continual pre-training is studied for domain adaptation where the new dataset comes from a new domain, which is referred to as continual domain-adaptive pre-training (DA-training)~\citep{gururangan2021demix,scialom2022finetuned,ke2023continual}. DAP-training methods can achieve better results by training LLMs with a large unlabeled domain corpus before end-task fine-tuning ~\citep{alsentzer2019publicly,Lee_2019,gururangan2020dont,ke2023adapting}. However, the effectiveness of this method is contingent upon the relevance of the pre-trained LLMs to the target domain and requires substantial domain-specific data to achieve optimal performance.

\paragraph{Alignment of Distributions}
Aligning the statistical attribution of the source and target domains to match their distributions has emerged as a principal method~\citep{peng2019moment,nguyenmeidine2020unsupervised}. 
The general way of aligning the distributions is by minimizing the distance between domains. The most used distance measures in domain adaptation are maximum mean discrepancy, Kullback-Leibler divergence, and contrastive domain discrepancy ~\citep{long2015learning,ganin2015unsupervised}.
The strength of this approach lies in its theoretical rigor and the potential for precise domain alignment. Nevertheless, the challenge of selecting appropriate distance metrics and the computational complexity of these calculations can pose significant obstacles. Compared with this, we adopt word-based statistical metrics to calculate the similarity between texts from different domains directly.

\paragraph{Adaptation Tuning}

LLMs may not capture sufficient knowledge for specific tasks or domains even when trained on vast amounts of general text data.
Adapting models to a smaller, domain-specific dataset can significantly improve their performance in that specific area.
Here are three primary methods for adapting LLMs: (1) Prompt engineering has shown its power to quickly adapt LLMs to unseen domains without updating the inner parameters. Prompts define unseen tasks with or without several illustrative examples to LLMs in natural language~\citep{bendavid2022pada,kojima2023large}. Continuous prompts are sequences of tokens attached with the input sentence that can be learned from the downstream dataset by prompt tuning ~\citep{Ye_2022,vu2022spot,razdaibiedina2023progressive}. \citet{Su_2022} demonstrate the transferability of continuous prompts in both cross-task as well as cross-model settings; (2) Adapter fine-tuning, such as Low-rank adapters~\citep{hu2021lora} and DyLora ~\citep{valipour2023dylora}, adds a small number of extra parameters to LLMs to enhance performance without major modifications; 
(3) Full fine-tuning is still the most fundamental and wildly used method to improve the model's adaptation performance. Instruction fine-tuning has proven to be highly successful in enhancing the model's adaptation capabilities~\citep{chung2022scaling,menick2022teaching,wei2022finetuned,huang2023language}. However, how to select suitable data to cultivate LLMs' adapting capacity and predict transferring results remain a problem.

\section{What and How does Word Influence Domain Adaptation?}
In this section, we explore how words can affect aspects of domain adaptation and their impact. We first hypothesize that datasets of varying learning difficulties affect model performance and investigate the influence of words on the learning difficulty of target domain datasets. We propose two indicators to reflect the learning difficulty of datasets: Compression Ratio and Abstraction Level.
Secondly, from the perspective of words, we propose two aspects that could affect domain adaptation: cross-domain overlap and word count. Finally, we hypothesize a linear relationship between these aspects and domain adaptation performance based on dataset learning difficulty. This hypothesis is tested in subsequent experiments.

\subsection{Word Influence On Target Domain Dataset Learning Difficulty}
    We assume that different datasets have varying levels of learning difficulty, and training models on datasets with low learning difficulty can lead to higher metric improvements on the test set. In contrast, the metric improvement is relatively small for datasets with high learning difficulty. To quantitatively assess the learning difficulty of datasets for the generative summarization task, we introduce two indicators: Compression Ratio and Abstraction Level. 

\paragraph{Compression Ratio}
    The Compression Ratio reflects the learning difficulty of a dataset in terms of form, which describes the degree of length reduction of the original text relative to the generated text. A higher compression Ratio indicates a more challenging dataset because the model needs to compress the content of the original documents to a greater extent, which places a higher demand on the model's text comprehension and information extraction capabilities. The Compression Ratio $\alpha$ for a dataset containing $n$ samples is represented as the average Compression Ratio across all samples and is calculated using the following formula:
    
\begin{equation}
    \alpha = \frac{1}{n} \sum\limits_{i=1}^{n}\frac{|D_i|}{|S_i|},
\end{equation}
    where $|D_i|$ and $|S_i|$ represent the word count of the $i$-th document and summary in the dataset, respectively.

\paragraph{Abstraction Level}
    Considering only the Compression Ratio cannot fully reflect the dataset's learning difficulty. For example, if the reference summaries in the dataset are verbatim excerpts of specific sentences from the source document, even though the Compression Ratio may be high, the model only needs to learn to copy parts of the document to achieve high performance. This is a straightforward extractive pattern and does not truly reflect the model's summarization capability. Therefore, we introduce Abstraction Level that reflects the learning difficulty of a dataset in terms of content, which we define as the reciprocal of the average ROUGE score between the original documents in the test set and the corresponding summaries, with the formula represented as follows:

\begin{equation}
    \beta = \frac{n}{\sum\limits_{i=1}^{n}{ROUGE_{(d_i,s_i)}}}
\end{equation}

    ROUGE is essentially a method for calculating overlap. We argue that the overlap between documents and summaries can, to some extent, represent the co-occurrence of knowledge within the dataset. A lower ROUGE value indicates lower content relevance between the document and the summary, making improving performance on that dataset. Hence, we use the reciprocal of Abstraction Level to reflect the learning difficulty of the dataset in terms of content.

\paragraph{Learning Difficulty Coefficient}
    According to the proposed two indicators influencing dataset learning difficulty, we define a dataset's learning difficulty coefficient $\lambda$ as the product of Compression Ratio and Abstraction Level, represented by the following formula:
    
\begin{equation}
\begin{aligned}
    &\lambda = \alpha \beta
\end{aligned}
\end{equation}

\subsection{Possible Impact Aspects Cross Different Domains Based on Words}
    We investigate the influence of words on domain adaptation performance. Due to the limitations of the original metric for summarization tasks, such as ROUGE, in reflecting how well a model generalizes across different domains, we introduce an evaluation metric suitable for assessing domain adaptation performance and explore the potential factors that might affect this metric. 

\paragraph{Summarization Gain}
    The commonly used evaluation metric in summarization tasks, ROUGE, reflects performance on a specific dataset, whereas the performance of domain adaptation is more evident in the change in absolute performance. Therefore, we employ the ROUGE gain as a fundamental measure of domain adaptation performance, with the formula as follows:

\begin{equation}
    Gain = ROUGE_{fine\text{-}tuned} - ROUGE_{base},
\end{equation}
    where $ROUGE_{base}$ represents the original model's ROUGE value calculated through direct inference on the test set, while $ROUGE_{fine\text{-}tuned}$ represents the ROUGE value obtained after the model has been fine-tuned.

\paragraph{Cross-domain Overlap}
    When considering performance adaptation across different domains, intuitively, they are more similar if there are more overlapping words between datasets from different domains. We assume this similarity can lead to performance improvement in domain adaptation. We propose cross-domain overlap to characterize the word-level overlap ratio between different domains. For source domain $S$ containing $n$ datasets and target domain $T$ containing $m$ datasets, the formula for cross-domain overlap is expressed as follows:

\begin{equation}
\begin{aligned}
    \gamma = \frac{1}{n \cdot m} \sum_{i=1}^{n} \sum_{j=1}^{m} \frac{\sum_{k=1}^{l} Count(w_k^{T_j}, S_i)}{|S_i|}
\end{aligned}
\end{equation}
where $S_i$ and $T_i$ represent the $i$-th dataset for the domain $S$ and $T$ respectively, $l$ is the total number of unique words in $T_j$, $w_k^{T_j}$ represents the $k$-th unique word in $T_j$, and $Count(w_k^{T_j},S_i)$ is the number of occurrences of word $w_i^{T_j}$ in $S_i$.

\paragraph{Word Count}
Word count refers to the total number of words across all samples in a dataset. Generally, the larger the number of training set samples within a specific range, the better the model performance. However, whether a higher total word count in all samples is also beneficial is worth exploring. Therefore, we consider word count as one aspect affecting the model's domain adaptation performance.

\subsection{How to influence?}
   We posit that the metrics of cross-domain overlap and Word Count notably influence domain adaptation performance, particularly when accounting for dataset learning difficulty.
    The higher the cross-domain overlap, the more similarity between the source domain's data and the target domain's data. Consequently, the model is more likely to leverage data from source domains to learn knowledge that is closer to the target domain, resulting in better adaptation to that domain. 
    Regarding Word Count, it is intuitive to assume that a larger training dataset size, and consequently a higher Word Count, leads to better model performance. However, whether this correlation consistently extends to domain adaptation performance remains an open question for investigation.

    \paragraph{LD-Gain} When dataset learning difficulty is factored into domain adaptation performance, for more challenging datasets, a model should require a smaller performance gain to achieve the same level of performance as when difficulty is not considered since the dataset's complexity impedes the model's ability to generalize across domains. Hence, we hypothesize that cross-domain overlap and Word Count are linearly related to the product of dataset learning difficulty and performance gain, which is referred to \textbf{LD-Gain}. The proposed hypotheses are as follows:
\begin{equation}
\begin{aligned}
     Hypothesis1: &\gamma \propto \lambda Gain = LD\text{-}Gain, \\
     Hypothesis2: &WC \propto \lambda Gain = LD\text{-}Gain,
\end{aligned}
\end{equation}
where $WC$ represents Word Count. The following experiments section will verify the two hypotheses.

\begin{table}
    \centering
    \resizebox{\linewidth}{!}{
    \begin{tabular}{lc rrrrrrr rrrrrrr}
    
\toprule
Dataset & Domain & \makecell{Training\\Set} & \makecell{Test\\Set} & \makecell{Document \\ Words (Avg.)} & \makecell{Summary \\ Words (Avg.)} \\
    
\midrule

\multirow{2}{*}{CNNDM} & \multirow{2}{*}{News} & 35,000 & 500 & 663 & 56 & \\
&& (287113) & (11490) & (781) & (56) & \\
\midrule
\multirow{2}{*}{PubMed} & \multirow{2}{*}{Science} & 35,000 & 500  & 1064 & 190 & \\
&& (119,924) & (6,658) & (3,049) & (202) & \\
\midrule
\multirow{2}{*}{SAMSum} & \multirow{2}{*}{Conversation} & 14,732 & 500 & 93 & 20 & \\
&& (14,732) & (819) & (92) & (20) & \\
\midrule
\multirow{2}{*}{WikiHow} & \multirow{2}{*}{General} & 35,000 & 500  & 523 & 52 & \\
&& (157,252) & (5,577) & (580) & (62) & \\

\bottomrule

    \end{tabular}
    }
    \caption{The statistics of datasets. The data in parentheses represent the total data from the original datasets.}
    \label{tab:statistics}
\end{table}

\begin{table*}[htp]
    \centering
    \resizebox{\linewidth}{!}{
    \begin{tabular}{ccccccccccc}
    
\toprule

\makecell{Training Set} & \makecell{Test Set} & \makecell{Compression\\Rate} & \makecell{Abstract\\Level} & \makecell{Learning Difficulty\\Coefficient} & \makecell{ROUGE\\Base} & \makecell{ROUGE\\Fine-tuned}& \makecell{ROUGE\\Improvement} & \makecell{Cross-domain\\Overlap} & $LD\text{-}Gain$ &  \\
\midrule

PubMed  & \multirow{3}{*}{CNNDM}  & \multirow{3}{*}{12.95} & \multirow{3}{*}{20.22} & \multirow{3}{*}{261.85} & \multirow{3}{*}{8.10} &7.84&-0.26& 2.35\%  & -68.08 \\
                        SAMSum  & &  &  &  & &10.23&2.13& 8.89\% & 557.74 \\
                        WikiHow & &  &  &  & &8.47&0.37& 4.47\% & 96.88 \\
\hdashline
CNNDM & \multirow{3}{*}{PubMed} & \multirow{3}{*}{7.08} & \multirow{3}{*}{13.13} & \multirow{3}{*}{92.96 } & \multirow{3}{*}{7.42} &10.65 & 3.28 &2.90\% &304.91 \\
                        SAMSum  &  &  &  & & &6.80  &-0.62 &3.51\% & -57.64 \\
                        WikiHow &  &  &  & & &6.42  &-1.00 &3.56\% & -92.96 \\
\hdashline
CNNDM & \multirow{3}{*}{SAMSum} & \multirow{3}{*}{4.86} & \multirow{3}{*}{16.76} & \multirow{3}{*}{81.45 } & \multirow{3}{*}{5.22} &6.36 &1.14 &1.62\% &92.85 \\
                        PubMed  &  &  &  &  &  & 4.61 &-0.61 &0.64\% &-49.68\\
                        WikiHow &  &  &  &  &  & 2.99 &-2.23 &1.26\% &-181.63 \\
\hdashline
CNNDM & \multirow{3}{*}{WikiHow} & \multirow{3}{*}{13.05} & \multirow{3}{*}{39.23} & \multirow{3}{*}{511.95 } & \multirow{3}{*}{4.562} & 5.075 &0.513 &3.30\% &262.63 \\
                        PubMed   &  &  &  &  &  &4.506 &-0.056 &2.25\% &-28.67\\
                        SAMSum   &  &  &  &  &  &5.39 &0.83 &7.53\% &424.92\\
 
\bottomrule
 
    \end{tabular}}
    \caption{Results of single-domain adaptation on the Llama-2 7B model.}
    \label{tab:single-7b}
\end{table*}

\begin{figure*}[ht]
	\centering
	\begin{minipage}[c]{0.32\textwidth}
		\centering
		\includegraphics[width=\textwidth]{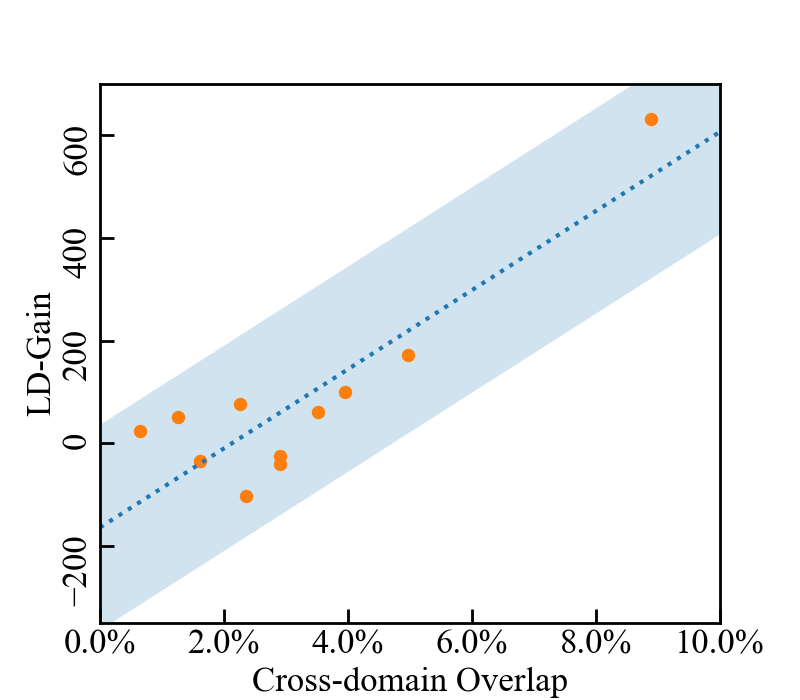}
		\subcaption{Bloom-1.1B}
		\label{fig:bloom1b}
	\end{minipage} 
	\begin{minipage}[c]{0.32\textwidth}
		\centering
		\includegraphics[width=\textwidth]{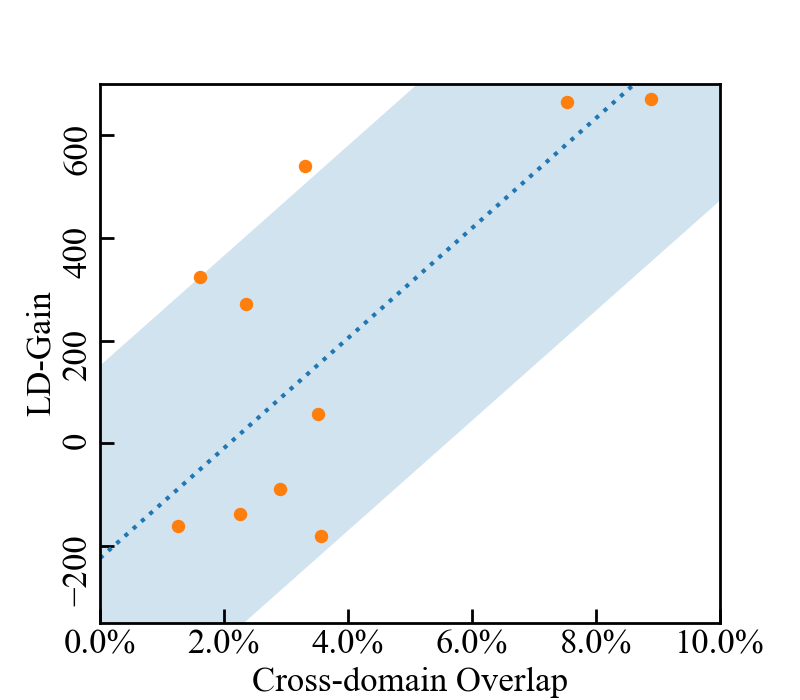}
		\subcaption{Bloom-3B}
		\label{fig:bloom3b}
	\end{minipage}
        \begin{minipage}[c]{0.32\textwidth}
		\centering
		\includegraphics[width=\textwidth]{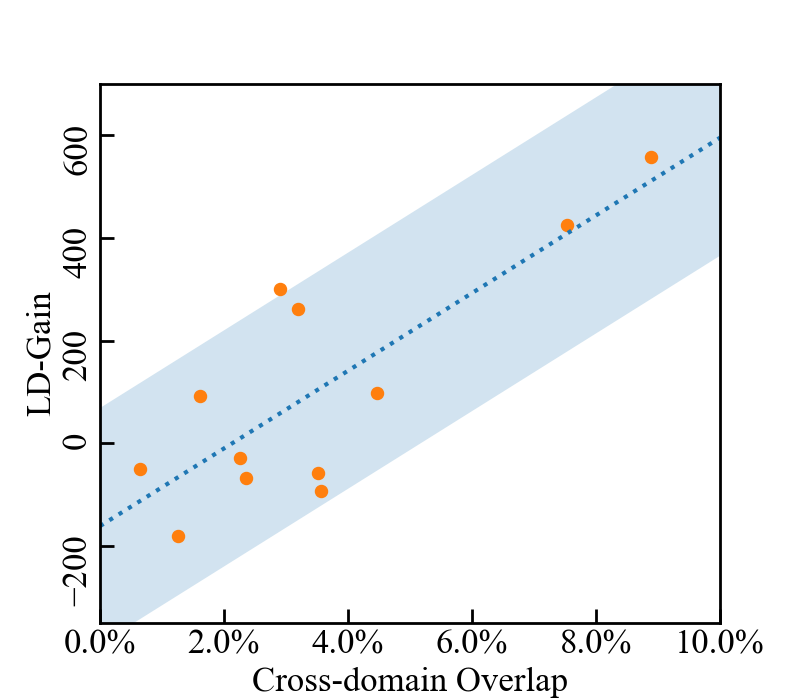}
		\subcaption{Llama2-7B}
		\label{fig:llama2}
	\end{minipage}
        \caption{Single-domain adaptation. The dotted line reflects a changing trend fitting the scatter data. The light-colored area represents a standard deviation of plus or minus.}
        \label{fig:transfer}
\end{figure*}

\begin{table*}[t]
\centering
\resizebox{\linewidth}{!}{
\begin{tabular}{l cccc ccc ccc cc}
\toprule
Model & CNNDM & PubMed & SAMSum & WikiHow & \makecell{Compression \\ Rate} & \makecell{Abstract \\ Level} & \makecell{Learning Difficulty\\Coefficient} & \makecell{ROUGE\\base}  & \makecell{ROUGE\\Fine-tuned} & \makecell{ROUGE\\Improvement} & \makecell{Cross-domain\\Overlap} & $LD\text{-}Gain$ \\

\midrule

\multirow{4}{*}{\bf Bloom-1B} & $\CIRCLE$ & $\Circle$ & $\Circle$ & $\Circle$ & 12.95 & 20.22& 261.85&3.72 &3.91
&0.19&3.13\% &49.75 \\
& $\Circle$ & $\CIRCLE$ & $\Circle$ & $\Circle$ & 7.08 & 13.13& 92.96&4.13 &4.65&0.52&3.18\% &48.34\\
& $\Circle$ & $\Circle$ & $\CIRCLE$ & $\Circle$ & 4.86 &16.76&81.45&1.75 &1.65&-0.1&1.05\% &-8.15\\
& $\Circle$ & $\Circle$ & $\Circle$ & $\CIRCLE$ & 13.05&39.23&511.95&2.51 &2.49&-0.02&2.60\% &-10.24\\

\midrule

\multirow{4}{*}{\bf Bloom-3B} & $\CIRCLE$ & $\Circle$ & $\Circle$ & $\Circle$ &  12.95 & 20.22& 261.85&4.06&5.48
&1.42&3.13\%&371.83 \\
& $\Circle$ & $\CIRCLE$ & $\Circle$ & $\Circle$ &  7.08 & 13.13& 92.96&4.84&6.87&2.03&3.18\%&188.71\\
& $\Circle$ & $\Circle$ & $\CIRCLE$ & $\Circle$ & 4.86 &16.76&81.45&1.81&2.17&0.36&1.05\%&29.32\\
& $\Circle$ & $\Circle$ & $\Circle$ & $\CIRCLE$ & 13.05&39.23&511.95&2.59&3&0.41&2.60\%&209.90\\

\bottomrule

\end{tabular}
}
\caption{
Multi-domain Adaptation. $\CIRCLE$ represents the test sets, while $\Circle$ constitutes the training sets together.
}
\label{tab:multi}
\end{table*}

\section{Experiments}
    We use four summarization datasets, CNN/Daily Mail~\citep{DBLP:conf/nips/HermannKGEKSB15}, PubMed~\citep{cohan-etal-2018-discourse}, SamSum~\citep{gliwa-etal-2019-samsum} and WikiHow~\citep{koupaee2018wikihow}, each originating from the news, science, conversation, and general domains respectively.
    Due to the significant differences in the number of samples across different datasets, we sample 35,000 samples from the CNNDM, PubMed, and WikiHow datasets for the training set while retaining the entire SAMSum dataset. We sample 500 test samples from all the test sets of these datasets.
    The detailed statistical data of the datasets are shown in Table~\ref{tab:statistics}.

    To verify the above two hypotheses, we configure different experimental setups for cross-domain overlap and Word Count. All experiments are based on the Bloom~\citep{Bloom} and Llama2~\citep{touvron2023llama} series of models. Due to resource constraints, different experiments use models of various sizes. The prompts used to train the Llama2 and Bloom models are presented in appendix~\ref{appendix_b}.

\subsection{Setup for Cross-domain Overlap}
Cross-domain overlap is calculated between the different source and target domains. Considering the potential impact of the number of source domains on the results, we set up two experiments: single-domain adaptation and multi-domain adaptation.

\paragraph{Single-domain Adaptation}
Single-domain adaptation refers to training on a single source domain and testing on different target domains.
We first test the basic performance of models on test sets across four domains and then fine-tune the models using training sets from the three domains, excluding the test set domain. For a model, this results in having models trained on three single-domain datasets, which are then tested for their performance on the test sets. Finally, we calculate the change in performance.
The Bloom-1.1B, Bloom-3B, and Llama-2 7B models are trained on 4 RTX 3090 GPUs in this experiment. For the Bloom-1.1B and 3B models, we conduct full-parameter fine-tuning for one epoch with a learning rate 2e-5 and a batch size of 4. For the Llama2-7B model, we use LoRA~\citep{hu2021lora} for fine-tuning over three epochs, with the other hyperparameters remaining unchanged.

\begin{figure}[ht]
	\centering
	\begin{minipage}[c]{0.45\linewidth}
		\centering
		\includegraphics[width=\textwidth]{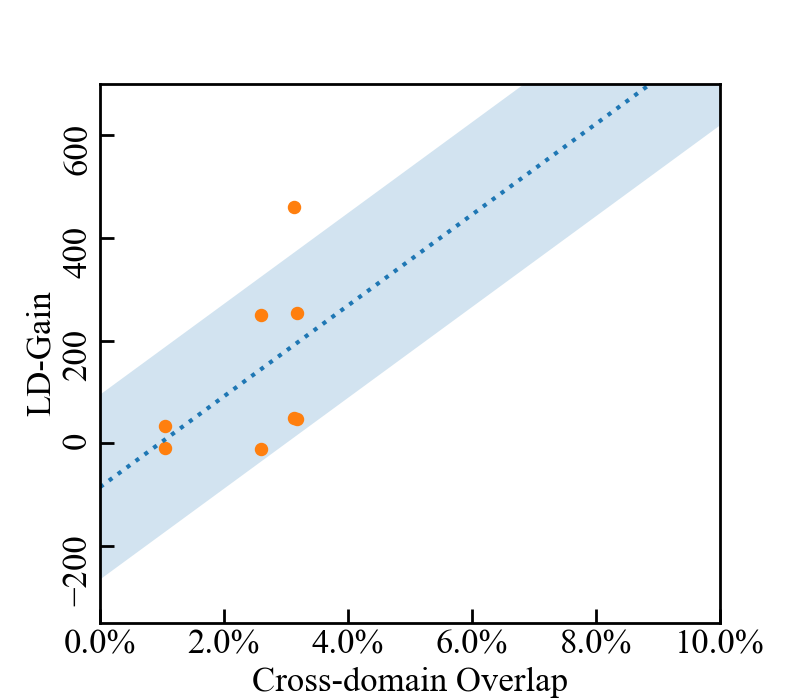}
		\subcaption{Multi-Domain}
		\label{fig:multi}
	\end{minipage} 
	\begin{minipage}[c]{0.45\linewidth}
		\centering
		\includegraphics[width=\textwidth]{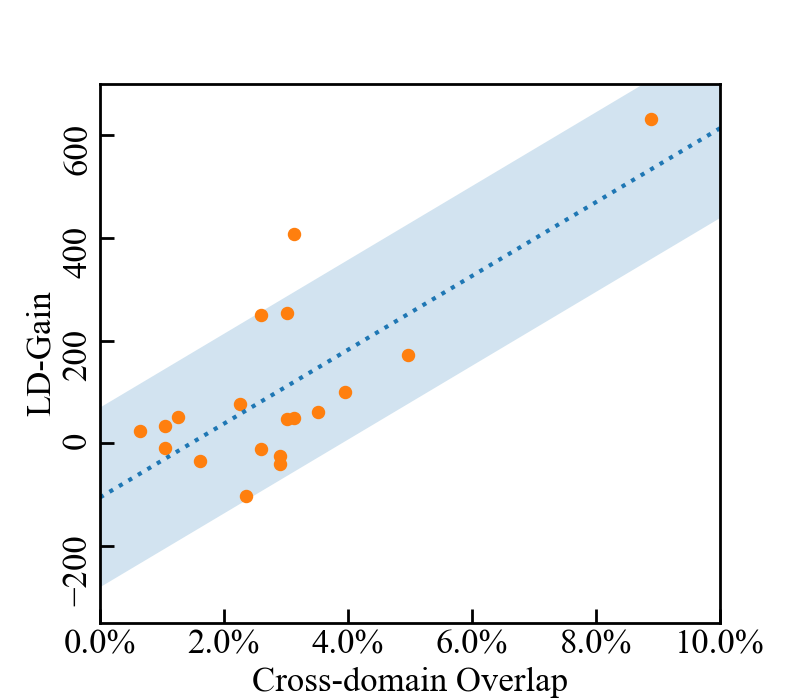}
		\subcaption{Mixed-Domain}
		\label{fig:single-multi}
	\end{minipage}
        \caption{The left is the result of multi-domain adaptation for Bloom-1.1B and 3B. The right contains the results of both single-domain and multi-domain adaptation.}
        \label{fig:multi-transfer}
\end{figure}

\paragraph{Multi-Domain Adaptation}
Multi-domain adaptation refers to training on multiple source domains. When one of the four domain datasets is selected as the test set, the datasets from the other three domains are combined to serve as the training set. In this experiment, we use the Bloom-1B and 3B models. The basic performance of the two models on the four domain test sets has already been tested in the single-domain adaptation, so we use the mixed dataset from the three domains, excluding the test set domain, for training to calculate the performance change.
The training hyperparameters are the same as those used in the single-domain adaptation.

\subsection{Setup for Word Count}
To minimize the interference of other factors in investigating the relationship between word count and domain adaptation, we use only one domain dataset, CNNDM, for training. Then, we test on the same domain's test set and the test sets of the other three domains to observe the impact of word count on the results. 
The CNNDM training set is evenly divided into ten chunks, each used for training in separate stages. Training is conducted in ten stages, starting with the first chunk as the training set. 
In subsequent stages, each phase uses the training set from the previous stage plus the next chunk, continuing until the final stage, where the entire dataset of the source domain is used for training. Each stage involves training for one epoch.

\section{Results and Analysis}
We analyze the results from different perspectives. Both single-domain and multi-domain experiment results supported our hypothesis that cross-domain overlap is linearly correlated with performance. Due to this finding, we could make a quantitative prediction about the transferred result of certain given training and evaluating datasets. 
On the other hand, we found that word count is unrelated to domain adaptation, which further emphasizes the significance of choosing high-quality data rather than a large amount of data.

To solidify the universality of our findings, we conduct a prediction study and it turns out that our hypothesis can successfully predict domain adaptation performance with cross-domain overlap. There is only a slight gap between our prediction data and observed data.

\subsection{Cross-domain overlap is linearly correlated with performance}
\paragraph{Single-Domain Adaptation}
    The results of the single-domain adaptation for Llama2-7B are shown in Table~\ref{tab:single-7b}, and the results of Bloom-1.1B and 3B are shown in Table~\ref{tab:single-1B} and Table~\ref{tab:single-3B} of Appendix~\ref{sec:appendix}. 
    The compression rate, abstraction level, and the learning difficulty coefficient calculated based on them only relate to the test set. Therefore, these indicators remain constant for the same test set, even if the training set changes. Similarly, the base ROUGE value of the model on the test set also remains unchanged.
    
Figure~\ref{fig:transfer} illustrates the relationship between cross-domain overlap and LD-Gain for three different models. It can be observed that when there is a low overlap between the target domain and the source domain, the model fails to generalize well to the target domain. 
Conversely, the model performance tends to exhibit significant improvement when there is high vocabulary overlap. 
We also utilize another similarity calculation method, BERTScore~\citep{Zhang2019BERTScoreET}, to replace ROUGE values in computing LD-Gain. The results for single-domain adaptation on bloom-3b and llama2-7b are depicted in the figure~\ref{fig:bertscore} of Appendix~\ref{results_bertscore}. We observe that the relationship between LD-Gain computed based on BERTScore and cross-domain overlap is similar to the results obtained with ROUGE-based calculations.
Based on this discovery, we believe that there exists a linear correlation between vocabulary overlap and model performance gain, factoring in dataset learning difficulty.

\paragraph{Multi-Domain Adaptation}
Table~\ref{tab:multi} presents the results obtained by training with multiple domain data and testing with a single domain. It can be observed that the ROUGE improvement of Bloom-1B on CNNDM is 0.19, which is smaller than 0.52 compared to the improvement on PubMed.
However, the learning difficulty coefficient for the CNNDM test set is higher at 261.85, exceeding the value of 92.96 for PubMed. Therefore, the LD-Gain of the model on CNNDM, adjusted by the learning difficulty coefficient, is higher than that of PubMed. 

The relationship between cross-domain overlap and LD-Gain for multi-domain adaptation is illustrated in Figure~\ref{fig:multi}.
We also combine the results of single-domain and multi-domain adaptation and plot them in a single graph, as shown in Figure~\ref{fig:single-multi}. It can be observed that as the cross-domain overlap increases, the model's actual gains gradually increase. This finding reveals that cross-domain overlap has a linear relationship with performance in domain adaptation.

\begin{table*}[ht]
    \centering
    \resizebox{\linewidth}{!}{
    \begin{tabular}{c c ccccc ccccc}
    
\toprule
& Test Set & Chunk0 & Chunk1 & Chunk2 & Chunk3 & Chunk4 & Chunk5 & Chunk6 & Chunk7 & Chunk8 & Chunk9 \\

\midrule
Word Count & & 2,608,429 & 2,594,348 & 2,591,597 & 2,573,224 & 2,641,309 & 2,558,078 & 2,582,793 & 2,594,890 & 2,607,471 & 2,614,820 \\

\midrule
\multirow{4}{*}{\makecell{Cross-domain \\Overlap}} & CNNDM & 7.37\% &
7.35\% &
7.39\% &
7.35\% &
7.28\% &
7.29\% &
7.33\% &
7.28\% &
7.25\% &
7.34\% \\
\cdashline{2-12}
& PubMed &2.93\% &
3.00\% &
2.94\% &
2.87\% &
2.91\% &
2.92\% &
2.92\% &
2.95\% &
2.93\% &
2.84\% \\
& SAMSum & 1.59\% &
1.56\% &
1.74\% &
1.57\% &
1.60\% &
1.55\% &
1.60\% &
1.60\% &
1.64\% &
1.76\%\\
& WikiHow & 3.58\% &
3.57\% &
3.64\% &
3.60\% &
3.56\% &
3.55\% &
3.54\% &
3.55\% &
3.48\% &
3.55\%\\

\midrule

\multirow{4}{*}{LD\text{-}Gain} & CNNDM & 851.00&
775.07&
971.45&
735.79&
960.98&
995.02&
1091.91&
811.73&
976.69&
1099.76\\
\cdashline{2-12}
& PubMed & 43.69 &
31.61 &
21.38 &
26.96 &
33.47 &
49.27 &
25.10 &
26.03 &
35.32 &
40.90 \\
& SAMSum & 6.52 &
0.81 &
13.03 &
-24.44 &
-1.63 &
-24.44 &
-20.36 &
-8.96 &
-38.28 &
-17.11 \\
& WikiHow &-102.39 &
-15.36 &
25.60 &
-66.55 &
-30.72 &
30.72 &
112.63 &
-35.84 &
20.48 &
-66.55  \\

\bottomrule
 
    \end{tabular}}
    \caption{Results of different chunks on test sets across various domains. The word count is increasing from Chunk0 to Chunk9.}
    \label{tab:chunk}
\end{table*}

\subsection{Word count is not related to performance}
    The word count results are presented in Table~\ref{tab:chunk}. It can be observed that the word count within a chunk is relatively similar, indicating that the word count can be controlled by increasing the number of chunks. On the other hand, the overlap within a domain does not vary significantly. It remains stable within a small range, allowing for the observation of the relationship between overlap and performance across different domains.
    
\begin{figure}[ht]
\centering
\includegraphics[width=0.45\textwidth]{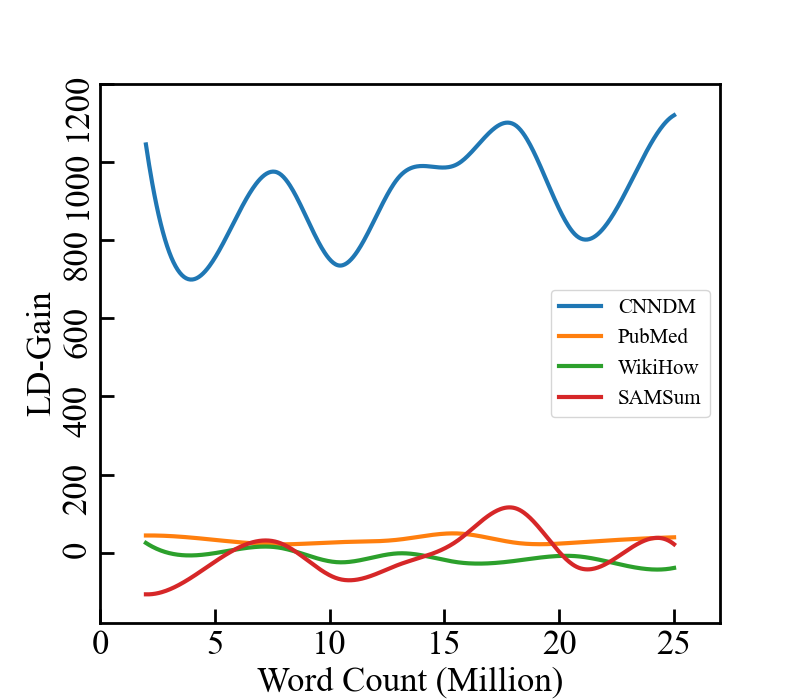}
\caption{
    Relationship between word count and LD-Gain. As word count increases, there is no significant trend in LD-Gain change.
}
\label{fig:slice-count}
\end{figure}

    The visual results from Figure~\ref{fig:slice-count} demonstrate that there is no clear upward or downward trend in model performance across different domains as the word count increases. Instead, there is oscillation within a specific range. The fluctuations are most prominent for the target domains CNNDM and SAMSum, which remain relatively stable within a specific range. Consequently, we conclude that the influence of word count on domain adaptation performance is unrelated. There are instances where performance may even decline with an increase in word count.

    Meanwhile, we also calculate the overlap of each chunk to observe whether there is still a linear relationship between the improvement of model performance and overlapping data at different domains. Although the cross-domain overlap is relatively similar within the same domain, a linear correlation exists between cross-domain overlap and LD-Gain across different domains. Specifically, the model exhibits higher actual gains as the cross-domain overlap increases.

\begin{figure}[ht]
\centering
\includegraphics[width=0.45\textwidth]{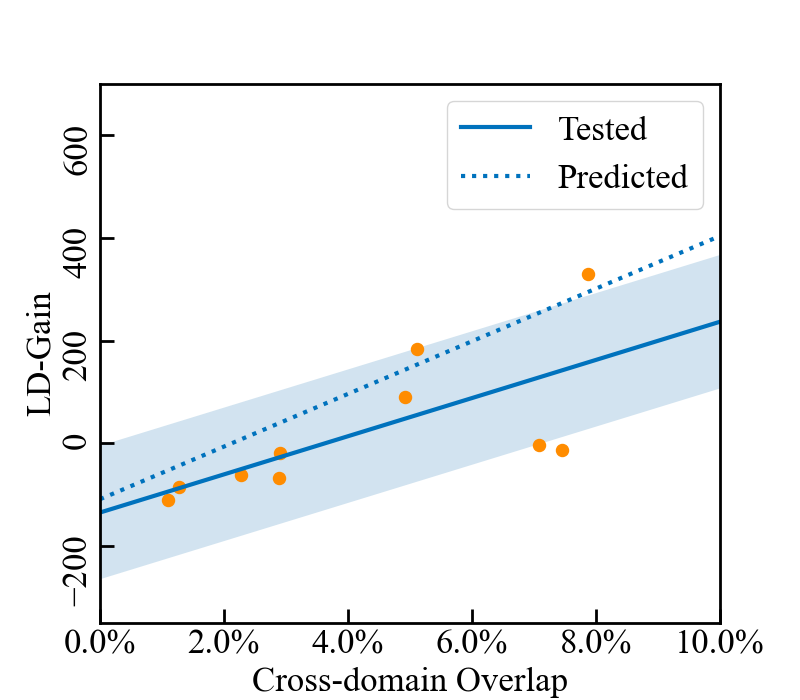}
\caption{
    The dotted line is the predicted line from the previous experiment, while the solid line is drawn with the testifying dataset.
}
\label{fig:prediction}
\end{figure}

\subsection{Predictability}
The preceding experiments have confirmed a linear correlation between the cross-domain overlap and performance, taking into account dataset learning difficulty. Based on this observation, we can extrapolate performance predictions for unknown domain datasets using the results from existing domain datasets. As depicted in Figure~\ref{fig:bloom1b}, a linear fit to the scatter plot of single-domain data yields a performance prediction trend for Bloom-1.1B in single-domain adaptation, as illustrated by the dashed line in Figure~\ref{fig:prediction}.

We re-sample 500 different examples, distinct from the previous datasets. Subsequently, we compute the compression ratio and abstraction level values for the four new test sets, obtaining the learning difficulty coefficient $\lambda$.  Based on the metrics from the training set used in the previous single-domain experiments, we calculate the cross-domain overlap $\gamma$ value. Ideally, we can use $\lambda$ and $\gamma$ to predict the model's performance gain on the new dataset, thereby obtaining the predicted ROUGE value with the formula as follows:
\begin{equation}
\begin{aligned}
ROUGE_{predicted}&=Gain+ROUGE_{base}, \\
Gain&=\frac{LD\text{-}Gain}{\lambda}, \\
LD\text{-}Gain&=\beta_0+\beta_1 \gamma,
\end{aligned}
\end{equation}
where $\beta$ represents the parameters of the fitting line under the existing data of the model.

We conduct experiments using the Bloom-1.1B model. The results of the new dataset, along with the fitted curve, are shown in Figure~\ref{fig:prediction} as scatter points and a solid line. It can be observed that the predicted fitted line and the actual experimental results' fitted line exhibit a similar trend, with a relatively small difference. This indicates that we can estimate the performance for unseen domain datasets based on the dataset's characteristics and existing performance results, thereby obtaining a rough performance expectation before actual inference.

\section{Conclusion}
\vspace{-0.1cm}
We investigate the impact of words on domain adaptation performance in summarization tasks. We propose two indicators to represent the learning difficulty from a dataset and introduce a performance evaluation method based on learning difficulty. 
We find that word overlap is an essential factor affecting domain adaptation and exhibits a linear correlation with model performance. However, the influence of word count on domain adaptation does not show a regular pattern.
We will investigate this phenomenon further in future work.

Furthermore, by predicting the performance of new domain data based on its cross-domain overlap with existing domains, it becomes possible to preemptively assess the model's suitability for specific domains without the need for extensive retraining or fine-tuning. This predictive capability can significantly streamline the adaptation of language models to new domains and save many resources, finally improving their practical utility in real-world applications. 

\section{Limitations}
The approaches of domain adaption mainly involve continual pre-training, alignment of distributions, and adaptation tuning. Our findings are related to the third one and, therefore, limited to discussing the influence factors of pre-training data, model parameters, and so on. 
For the adaptation tuning method, our paper focuses on exploring word impact on domain adaptation. 
Other factors, such as the quality of summarization training data, instruction diversity, and quality, are out of our consideration and may bring additional noise. 

Due to resource constraints, this paper employed LoRA fine-tuning for a 7B model without investigating the effects of different fine-tuning methods on domain adaptation.
In future work, we will explore in more detail the impact of dataset quality and different training methods on domain adaptation.

\section*{Acknowledgements}
We appreciate Xiaochen Liu for providing the initial inspiration for this work.
This work was supported by the Joint Funds of National Natural Science Foundation of China (No. U21B2009), Major Research Plan of the National Natural Science Foundation of China (Grant No. 92370110).

\bibliography{anthology,custom}
\bibliographystyle{acl_natbib}

\clearpage

\begin{appendix}

\section{The training prompts}
\label{appendix_b}
The prompt used to train the Llama2 model.
\vspace{-3mm}
\begin{center}
\fcolorbox{black}{gray!10}{
\parbox{2\linewidth}{
\small \texttt{<s> [INST] <<SYS>> You are a helpful, respectful and honest assistant. Always answer as helpfully as possible while being safe. Your answers should not include any harmful, unethical, racist, sexist, toxic, dangerous, or illegal content. Please ensure that your responses are socially unbiased and positive in nature.
\newline
\newline
If a question does not make any sense, or is not factually coherent, explain why instead of answering something not correct. If you don't know the answer to a question, please don't share false information.
<</SYS>>
\newline
\newline
Summarize the following paragraph
\newline
<Dodument>
\newline
[/INST]}}}
\end{center}

\vspace{3mm}
The prompt used to train the Bloom model.
\vspace{-3mm}
\begin{center}
\fcolorbox{black}{gray!10}{
\parbox{2\linewidth}{
\small \texttt{<s>A chat between a curious user and an artificial intelligence assistant. The assistant gives helpful, detailed, and polite answers to the user's questions.
summarize the following paragraph
<Document>}}}
\end{center}

\section{Results of Bloom-1.1B and 3B}
\label{sec:appendix}
\begin{table*}[ht]
    \centering
    \resizebox{\textwidth}{!}{
    \begin{tabular}{ccccccccccc}
    
\toprule

\makecell{Training Set} & \makecell{Test Set} & \makecell{Compression\\Rate} & \makecell{Abstract\\Level} & \makecell{Learning Difficulty\\Coefficient} & \makecell{ROUGE\\Base} & \makecell{ROUGE\\Fine-tuned}& \makecell{ROUGE\\Improvement} & \makecell{Cross-domain\\Overlap} & $LD\text{-}Gain$ &  \\
\midrule

                        PubMed  & \multirow{3}{*}{CNNDM}  & \multirow{3}{*}{12.95} & \multirow{3}{*}{20.22} & \multirow{3}{*}{261.85} & \multirow{3}{*}{3.72} &3.33&-0.39& 2.35\%  & -102.12 \\
                        SAMSum  & &  &  &  & &6.13&2.41& 8.89\% & 631.06 \\
                        WikiHow & &  &  &  & &4.38&0.66& 4.47\% & 172.82 \\
\hdashline
CNNDM & \multirow{3}{*}{PubMed} & \multirow{3}{*}{7.08} & \multirow{3}{*}{13.13} & \multirow{3}{*}{92.96 } & \multirow{3}{*}{4.13} &3.87 & -0.26 &2.90\% &-24.17 \\
                        SAMSum  &  &  &  & & &4.78  &0.65 &3.51\% & 60.42 \\
                        WikiHow &  &  &  & & &5.2  &1.07 &3.96\% & 99.47 \\
\hdashline
CNNDM & \multirow{3}{*}{SAMSum} & \multirow{3}{*}{4.86} & \multirow{3}{*}{16.76} & \multirow{3}{*}{81.45 } & \multirow{3}{*}{1.75} &1.32 &-0.43 &1.62\% &-35.03 \\
                        PubMed  &  &  &  &  &  & 2.04 &0.29 &0.64\% &23.62\\
                        WikiHow &  &  &  &  &  & 2.37 &0.62 &1.26\% &50.50 \\
\hdashline
CNNDM & \multirow{3}{*}{WikiHow} & \multirow{3}{*}{13.05} & \multirow{3}{*}{39.23} & \multirow{3}{*}{511.95 } & \multirow{3}{*}{2.51} & 2.43 &-0.08 &3.30\% &-40.96 \\
                        PubMed   &  &  &  &  &  &2.66 &0.15 &2.25\% &76.79\\
                        SAMSum   &  &  &  &  &  &2.46 &-0.05 &7.53\% &-25.6\\
 
\bottomrule
 
    \end{tabular}
    }

    \caption{Results of single-domain adaptation on the Bloom-1.1B model.}
    \label{tab:single-1B}
\end{table*}

\begin{table*}[ht]
    \centering
    \resizebox{\linewidth}{!}{
    \begin{tabular}{ccccccccccc}
    
\toprule

\makecell{Training Set} & \makecell{Test Set} & \makecell{Compression\\Rate} & \makecell{Abstract\\Level} & \makecell{Learning Difficulty\\Coefficient} & \makecell{ROUGE\\Base} & \makecell{ROUGE\\Fine-tuned}& \makecell{ROUGE\\Improvement} & \makecell{Cross-domain\\Overlap} & $LD\text{-}Gain$ &  \\
\midrule

PubMed  & \multirow{3}{*}{CNNDM}  & \multirow{3}{*}{12.95} & \multirow{3}{*}{20.22} & \multirow{3}{*}{261.85} & \multirow{3}{*}{6.348} &7.38&1.03& 2.35\%  & 269.71 \\
                        SAMSum  & &  &  &  & &8.91&2.56& 8.89\% & 670.34 \\
                        WikiHow & &  &  &  & &5.61&-0.74& 4.47\% & -193.77 \\
\hdashline
CNNDM & \multirow{3}{*}{PubMed} & \multirow{3}{*}{7.08} & \multirow{3}{*}{13.13} & \multirow{3}{*}{92.96 } & \multirow{3}{*}{8.60} &7.64 & -0.96 &2.90\% &-89.24 \\
                        SAMSum  &  &  &  & & &9.20  &0.61 &3.51\% & 56.71 \\
                        WikiHow &  &  &  & & &6.66  &-1.94 &3.56\% & -180.34 \\
\hdashline
CNNDM & \multirow{3}{*}{SAMSum} & \multirow{3}{*}{4.86} & \multirow{3}{*}{16.76} & \multirow{3}{*}{81.45 } & \multirow{3}{*}{5.19} &9.16 &3.97 &1.62\% &323.36 \\
                        PubMed  &  &  &  &  &  & 3.22 &-1.97 &0.64\% &-160.46\\
                        WikiHow &  &  &  &  &  & 0.66 &-4.53 &1.26\% &-368.97 \\
\hdashline
CNNDM & \multirow{3}{*}{WikiHow} & \multirow{3}{*}{13.05} & \multirow{3}{*}{39.23} & \multirow{3}{*}{511.95 } & \multirow{3}{*}{3.709} & 4.77 &1.06 &3.30\% &542.67 \\
                        PubMed   &  &  &  &  &  &3.44 &-0.27 &2.25\% &-138.23\\
                        SAMSum   &  &  &  &  &  &5.00 &1.30 &7.53\% &665.54\\
 
\bottomrule
 
    \end{tabular}}
    \caption{Results of single-domain adaptation on the Bloom-3B model.}
    \label{tab:single-3B}
\end{table*}

\section{Results of BERTScore}
\label{results_bertscore}
\begin{figure*}[ht]
	\centering
	\begin{minipage}[c]{0.45\textwidth}
		\centering
		\includegraphics[width=\textwidth]{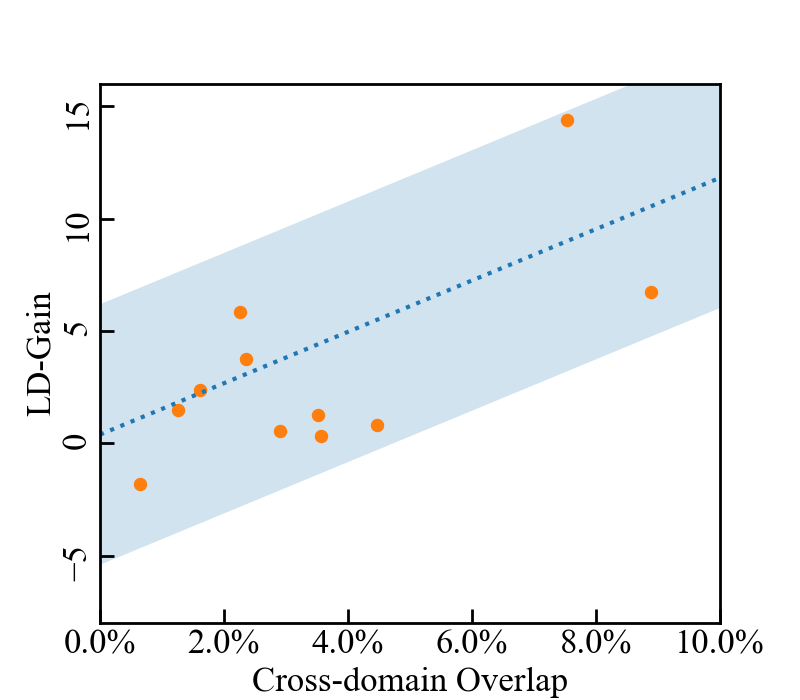}
		\subcaption{Bloom-3B}
		\label{fig:bertscore1}
	\end{minipage} 
	\begin{minipage}[c]{0.45\textwidth}
		\centering
		\includegraphics[width=\textwidth]{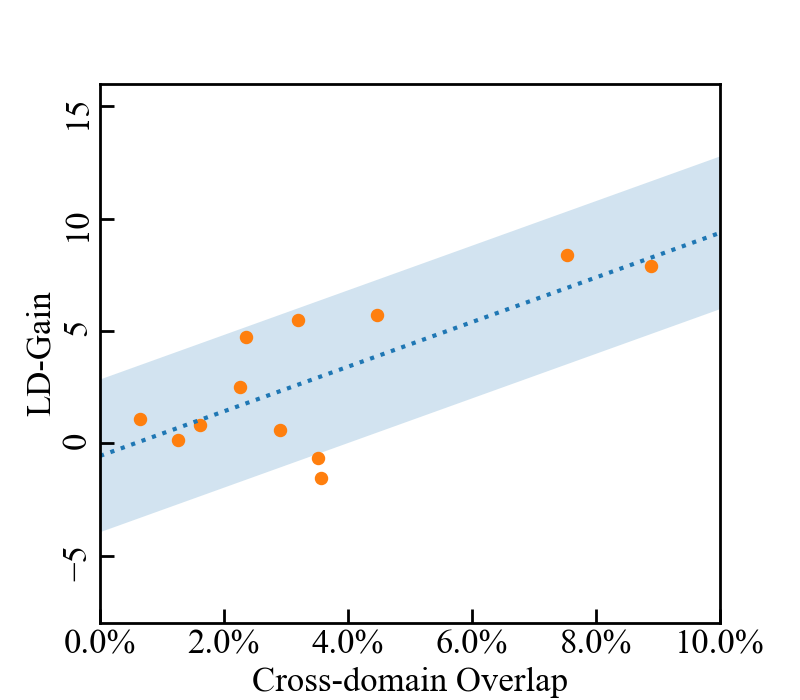}
		\subcaption{Llama2-7B}
		\label{fig:bertscore2}
	\end{minipage}
        \caption{Results of single-domain adaptation calculated by BERTScore.}
        \label{fig:bertscore}
\end{figure*}

\end{appendix}
\end{document}